\documentclass[letterpaper, 10 pt, conference]{ieeeconf}
\usepackage{amsmath,amsfonts}
\usepackage{algorithmic}
\usepackage{array}
\usepackage[caption=false,font=normalsize,labelfont=sf,textfont=sf]{subfig}
\usepackage{textcomp}
\usepackage{stfloats}
\usepackage{url}
\usepackage{verbatim}
\usepackage{graphicx}
\usepackage{microtype}
\def\BibTeX{{\rm B\kern-.05em{\sc i\kern-.025em b}\kern-.08em
    T\kern-.1667em\lower.7ex\hbox{E}\kern-.125emX}}
\usepackage{balance}

\usepackage{pifont}
\newcommand{\crossmark}{\ding{55}}  %
\usepackage{tabularx}
\usepackage{multirow}
\usepackage{booktabs}
\usepackage[table,xcdraw]{xcolor}
\usepackage{xspace}

\usepackage{svg}
\usepackage{tcolorbox}
\usepackage{listings}
\usepackage{makecell}

\definecolor{DarkPurple}{RGB}{103, 78, 167}

\newcommand{\empstar}{\ding{73}}
\newcommand{\fulstar}{\ding{72}}

\def\secref#1{Sec.~\ref{#1}}
\def\figref#1{Fig.~\ref{#1}}
\def\tabref#1{Tab.~\ref{#1}}
\def\eqref#1{Eq.~(\ref{#1})}

\makeatletter
\DeclareRobustCommand\onedot{\futurelet\@let@token\@onedot}
\def\@onedot{\ifx\@let@token.\else.\null\fi\xspace}

\def\ie{\emph{i.e}\onedot}

\def\etal{\emph{et al}\onedot}
\makeatother

\IEEEoverridecommandlockouts

\begin{document}

\title{AI-driven Dispensing of Coral Reseeding Devices for\\Broad-scale Restoration of the Great Barrier Reef}

\author{Scarlett Raine$^{1}$, Emilio Olivastri$^{1}$, Benjamin Moshirian$^{2}$ and Tobias Fischer$^{1}$%
\thanks{$^{1}$Scarlett Raine, Emilio Olivastri and Tobias Fischer are with the QUT Centre for Robotics and School of Electrical Engineering and Robotics, Queensland University of Technology, Brisbane, Australia.}%
\thanks{$^{2}$Benjamin Moshirian is with the Australian Institute of Marine Science, Townsville, Australia.}
\thanks{(\textit{Corresponding author: Scarlett Raine.} {\tt\footnotesize sg.raine@qut.edu.au})} %
\thanks{S.R., E.O.~and T.F.~acknowledge continued support from the Queensland University of Technology (QUT) through the Centre for Robotics. T.F.~acknowledges funding from an Australian Research Council (ARC) Discovery Early Career Researcher Award (DECRA) Fellowship DE240100149. S.R., E.O., B.M.~and T.F.~ acknowledge support from the Reef Restoration and Adaptation Program (RRAP) which is funded by a partnership between the Australian Government’s Reef Trust and the Great Barrier Reef Foundation.}%
}

\maketitle

\begin{abstract}
Coral reefs are on the brink of collapse, with climate change, ocean acidification, and pollution leading to a projected 70-90\% loss of coral species within the next decade. Reef restoration is crucial, but its success hinges on introducing automation to upscale efforts. In this work, we present a highly configurable AI pipeline for the real-time deployment of coral reseeding devices. The pipeline consists of three core components: (i) the image labeling scheme, designed to address data availability and reduce the cost of expert labeling; (ii) the classifier which  performs automated analysis of underwater imagery, at the image or patch-level, while also enabling quantitative coral coverage estimation; and (iii) the decision-making module that determines whether deployment should occur based on the classifier’s analysis. By reducing reliance on manual experts, our proposed pipeline increases operational range and efficiency of reef restoration. We validate the proposed pipeline at five sites across the Great Barrier Reef, benchmarking its performance against annotations from expert marine scientists. The pipeline achieves 77.8\% deployment accuracy, 89.1\% accuracy for sub-image patch classification, and real-time model inference at 5.5 frames per second on a Jetson Orin. To address the limited availability of labeled data in this domain and encourage further research, we publicly release a comprehensive, annotated dataset of substrate imagery from the surveyed sites.
\end{abstract}

\section{Introduction}
\label{sec:intro}

Coral reefs are critical marine ecosystems that support over 25\% of marine species and provide major economic benefits, especially through tourism and fisheries. However, ocean warming and repeated bleaching events are accelerating biodiversity decline and ecosystem degradation~\cite{henley2024highest}. Despite their ecological and economic importance, restoration initiatives are hindered by highly labor-intensive and expensive methodologies~\cite{hughes2023principles}. Coral gardening is a reef restoration technique that involves collecting small fragments of coral (asexual fragmentation), which are later planted on degraded reefs. The main limitations of this technique are low coral genetic diversity and the need for intensive, specialized, manual labor, which directly limits scalability~\cite{randall2020sexual}.

\begin{figure}[t]
 \centering
 \includegraphics[width=0.95\columnwidth,clip,trim=4.5cm 3.0cm 8.5cm 2.5cm]{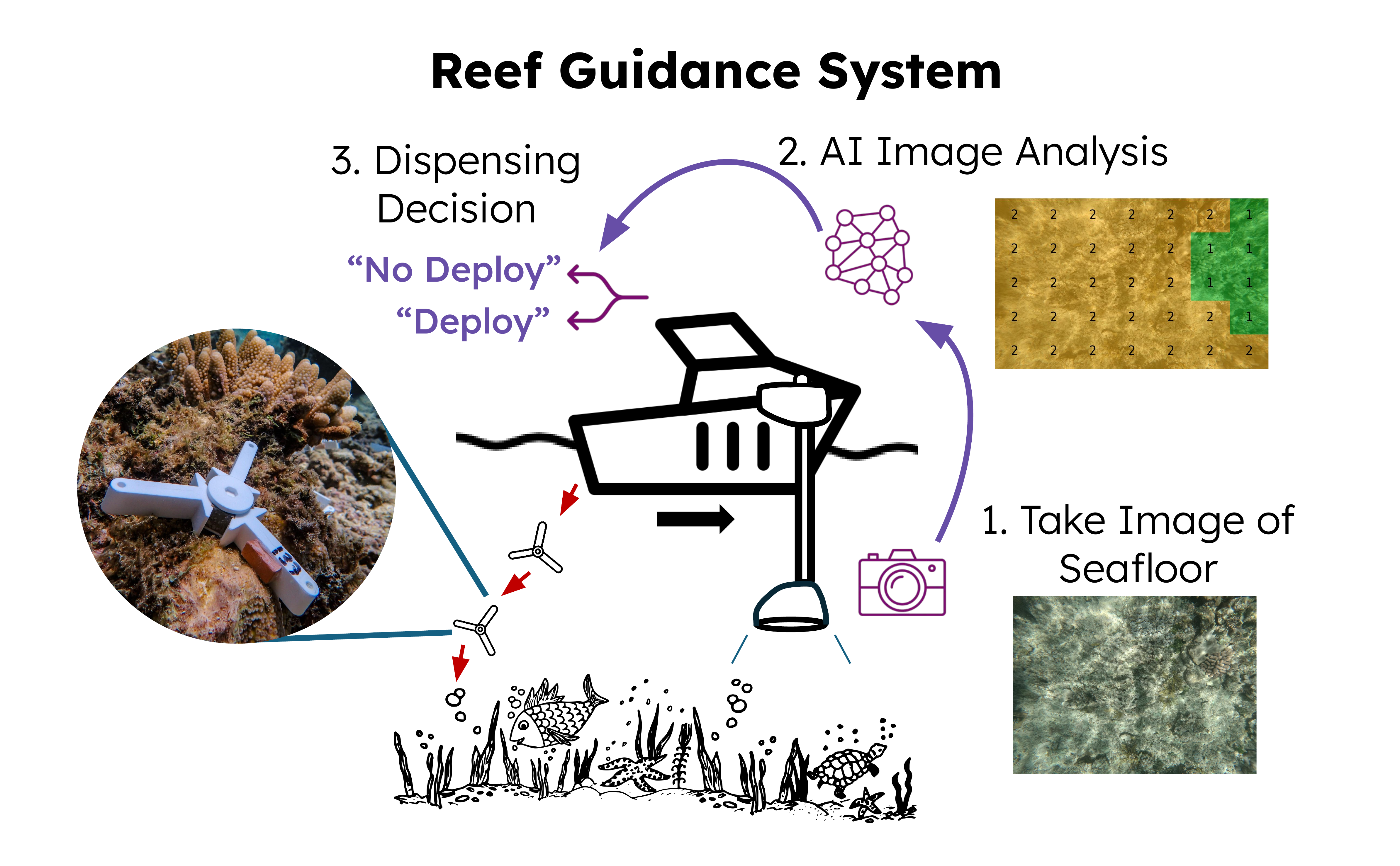}
 \caption{The Reef Guidance System incorporates a camera system which provides top-down, wide field of view, high resolution imagery of the seafloor; on-board compute to run the Artificial Intelligence (AI) model in real-time; a weakly supervised model training framework; and automated deployment of coral reseeding devices based on model decisions.}
 \label{fig:frontpage}
\end{figure}

Coral reseeding is an alternative technique for reef restoration, which involves collecting or breeding coral spawn (sexual reproduction), using coral aquaculture to grow corals until they reach a more resilient development stage, and then delivering the corals to regions of the reef which require restoration but will support their survival and growth~\cite{banaszak2023applying}. The corals are housed in ceramic `devices' which protect them during deployment and provide a stable foundation for their establishment (Fig.~\ref{fig:frontpage}). Unlike coral gardening, coral reseeding enhances genetic diversity and fosters populations with greater resilience~\cite{tsai2026automated}.

However, coral reseeding is currently still mostly performed manually by divers and snorkelers. As a step towards automation, experts are transitioning to release coral devices from a vessel guided by visual feedback from cameras. %
This human-dependent workflow restricts scalability and substantially reduces the efficiency of restoration operations. As restoration efforts aim to deliver thousands and eventually tens of thousands of corals to reefs, the extended use of technology and automation to support deployment becomes paramount~\cite{severati2024autospawner, gibbs2021technology}. 

While recent advances in AI-driven monitoring have enhanced reef assessment~\cite{raine2026ai} and facilitated more targeted restoration strategies~\cite{belcher2023demystifying, raine2024reducing, huang2024research}, the primary bottleneck lies in the automation of intervention operations~\cite{severati2024autospawner, gibbs2021technology}. This bottleneck stems from the substantial hardware, operational, and logistical demands of field-ready systems, the lack of standardized annotated datasets for these tasks, as well as the stochastic and highly variable environmental conditions that challenge reliable perception in real-world marine settings.

Furthermore, automated coral reseeding using AI is complicated by the challenge of visually determining suitable reef substrate, a task that has traditionally required a highly trained marine ecologist. In this work, we propose using robotics, computer vision and artificial intelligence in combination with specialized expert knowledge to enable automated broad-scale deployment of coral devices for the first time.  We present the Reef Guidance System (RGS; see Fig.~\ref{fig:frontpage}), which is an AI model that classifies the reef substrate to determine suitable locations for coral devices to be deployed. It runs on an edge camera unit which informs when to drop coral devices that are carried in the vessel. The RGS performs automated substrate analysis and deployment decision-making in real-time at depths up to 10m. 

Our paper presents the following contributions:
\begin{enumerate}
    \item We propose a flexible AI pipeline for automated coral reseeding in real-time, integrating labeling, classification, and deployment modules.
    \item We demonstrate the effectiveness of the pipeline for real-world reef restoration across five Great Barrier Reef sites, achieving 77.8\% accuracy compared to expert ecologists on challenging seafloor imagery.
    \item The pipeline is designed to accommodate different operational requirements, including accuracy, inference speed, interpretability, vessel configuration, and availability of domain experts.
\end{enumerate}

This work forms part of the large-scale Reef Restoration and Adaptation Program (RRAP), a global leader in coral reef restoration and adaptation research that develops and deploys innovative solutions to protect, restore, and build reef resilience.
We make our code and data available to foster future research on substrate classification for reef restoration\footnote{Code will be made available upon acceptance.}.%
\section{Related Work}
\label{sec:relatedwork}

Recent advances in AI and robotics enable the collection and automatic analysis of underwater imagery, increasing the scale and efficiency of marine surveys~\cite{ditria2022artificial, kumar2025harnessing}. However, the problem of automated coral reseeding poses a set of unique challenges: the concept of seafloor suitability for coral reseeding~\cite{darkhal2025optimizing} is more open to interpretation than the presence/absence of a species, and many different visual conditions could result in a positive deployment decision.  This section discusses prior approaches for underwater scene understanding and recent advances in AI-driven reef restoration. 

\subsection{AI-driven Underwater Image Analysis}

Prior approaches have leveraged deep learning to automate the analysis of underwater imagery~\cite{belcher2023demystifying, raine2024reducing}.  These works have performed multi-species coral segmentation~\cite{zheng2024coralscop, raine2022point}, with Raine \etal~\cite{raine2024human} proposing a human-in-the-loop annotation scheme, where human experts work in conjunction with an AI model to efficiently label coral imagery. There have also been advances in monitoring of seagrass meadows through image analysis~\cite{raine2024image, noman2023improving} and 3D habitat mapping~\cite{zhang2024point}. 

There have been numerous approaches for classification of coral imagery~\cite{chen2021new, gonzalez2020monitoring, wyatt2022using}. In~\cite{jagadeesh2025coral}, the authors use a combination of environmental factors (\ie turbidity, cyclone rates, temperatures) to predict when coral bleaching will occur. However, despite an increase in these approaches in the literature~\cite{raine2026ai}, the uptake of AI-driven methods for real-world field deployment has been limited in the field of marine science: Belcher \etal~\cite{belcher2023demystifying} present a simplified guide and case study for underwater species detection and classification, with the aim of making these tools more approachable to non-AI experts. Hanke \etal~\cite{hanke2025out} use computer vision techniques to extract sections of interest of underwater videos which capture the movement of the wildlife of interest, however they highlight the need for collaboration between marine scientists and AI researchers in applying these technologies effectively in marine conservation.  

To improve the robustness and stability of image classifiers, the authors in~\cite{wyatt2022using} propose an ensemble strategy for measuring the reliability of coral image classification in new or unseen environments, and~\cite{wyatt2026signal} explore the extent of bias and error in an AI pipeline for monitoring macroalgal forests in benthic imagery. Although these approaches take steps towards field-ready systems for monitoring marine ecosystems, there are fewer approaches for automated substrate analysis~\cite{jackett2023benthic, huang2024research}, and these approaches have not been leveraged for the purpose of determining suitable substrates for coral reseeding or are not designed for real time inference~\cite{loureiro2024survey}.

\subsection{AI-driven Reef Restoration}

While AI has been used extensively for monitoring coral reef ecosystems, there are limited approaches which perform active restoration. Morand \etal~\cite{morand2022identifying} used AI to monitor the growth of corals on restoration frames, while Tsai \etal~\cite{tsai2026automated} used AI to monitor the fertilization rates and coral stages of growth during coral rearing in an aquaculture facility. Neither of these works cover the task of coral reseeding. Aguzzi \etal~\cite{aguzzi2024new} highlight the potential use of Remotely Operated Vehicles and Autonomous Underwater Vehicles for collection, transplanting and reintroduction of species in deep-sea reef restoration.  The LarvalBot AUV and LarvalBoat ASV~\cite{dunbabin2020uncrewed} are automated coral larvae deployment systems, however their size and power requirements are not suitable for carrying and deploying the ceramic coral devices used in this work.

\begin{figure*}[t]
\centering
\includegraphics[width=0.95\linewidth,clip,trim=0cm 4cm 0cm 4cm]{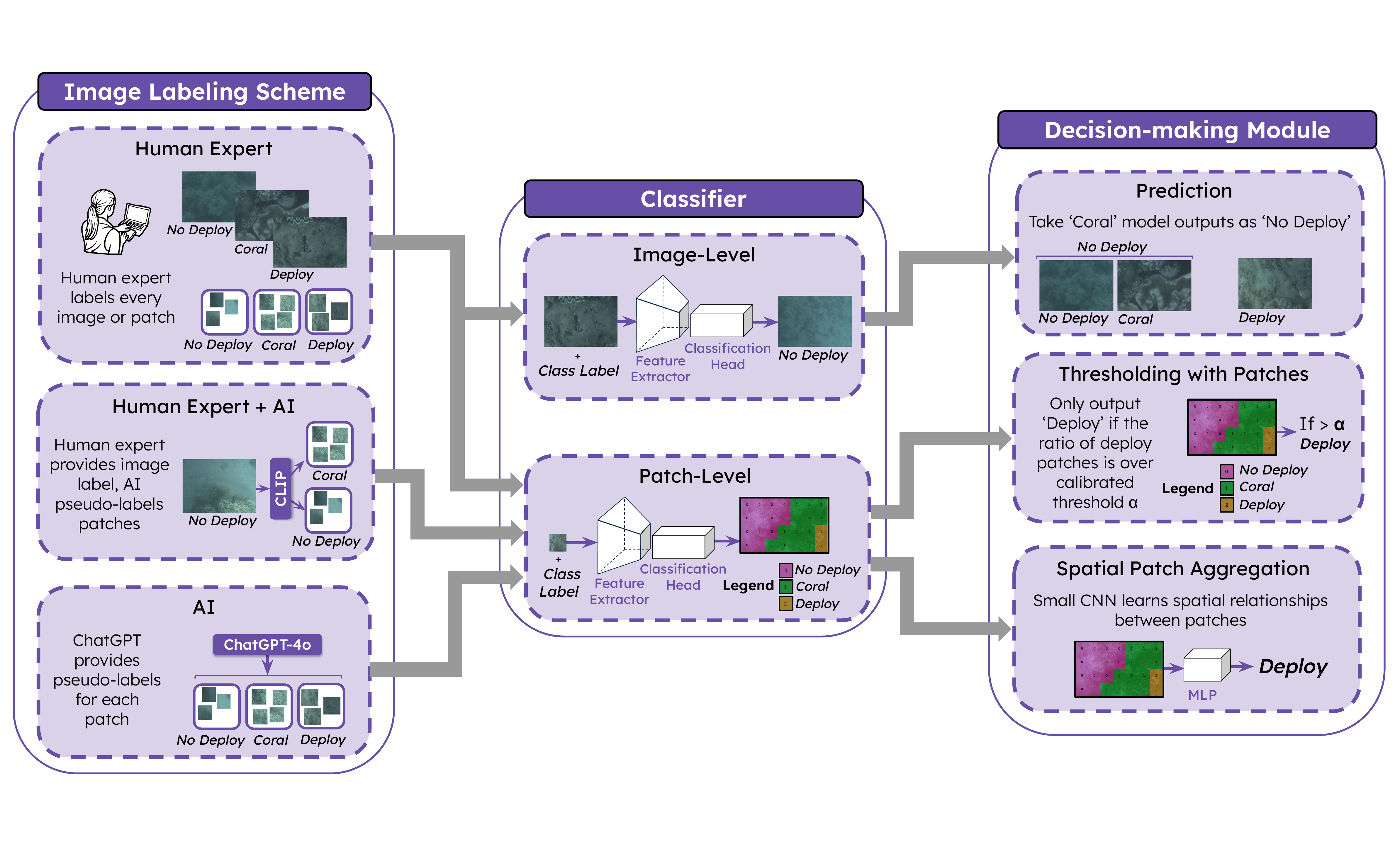}
\caption{\textbf{Proposed Pipeline Schematic}. We propose a flexible pipeline where different variations of our approach can be selected to best suit the operational requirements of the reef restoration program.  Left: the user can select an Image Labeling Scheme from three options based on whether a domain expert is available to annotate data examples of new locations and environmental characteristics. Each scheme has differing levels of human expert input.  Center: we propose two classifiers, image-level and patch-level, to provide flexibility in terms of the level of interpretability of model outputs as well as the configuration of the boat during coral reseeding (\ie whether the devices can be deployed from one side or both sides of the vessel simultaneously; the location of the deployment hardware on the vessel).  Right: we propose three alternatives for the decision-making module and evaluate the performance in terms of accuracy and inference speed.}
\label{fig:pipeline}
\end{figure*}

\section{Operational Considerations for Coral Reseeding}
\label{sec:operational}
The objective of this work is to propose a highly configurable pipeline for automated coral deployment, designed to accommodate the diverse operational constraints faced by organizations engaged in coral reseeding. Here we discuss the five main factors to consider when designing a field-ready pipeline for coral reseeding:
\begin{enumerate}
    \item \textbf{Accuracy}: The model must reliably determine when to deploy a coral device. For practical viability, the model's deployment locations should align with those chosen by human experts, while maintaining a comparable deployment rate to theirs.
    \item \textbf{Inference Speed}: In a typical deployment scenario, as illustrated in \figref{fig:frontpage}, cameras are mounted on a moving vessel operating at around 1 m/s. Real-time operation requires that deployment decisions must be completed before the vessel moves beyond the observed location. Consequently, inference speed directly constrains vessel velocity, while operational requirements impose strict upper bounds on allowable decision latency.
    \item \textbf{Interpretability}: Since the pipeline supports marine scientists, the decisions must be easily interpretable to non-AI experts. Interpretability enables operators to validate outputs, diagnose errors, and assess system limitations.   
    \item \textbf{Flexibility}: The solution must generalize across different hardware configurations, including variations in camera count, placement, and vessel setup. 
    \item \textbf{Expert Data Labeling}: Limited availability of annotated data presents a major challenge. Marine scientists are required to produce reliable labels, making the process costly, time-consuming, and prone to human error. Reducing annotation burden is therefore an important system requirement.
\end{enumerate}

\section{Method}
\label{sec:method}
In this section, we present the proposed pipeline for automated coral reseeding. This pipeline has three principal components: the image labeling scheme used for creating the training dataset, the classifier for image analysis, and the decision-making module. We adopt MobileNetv3 as the backbone classifier due to its high inference speed on low-power embedded devices, such as the Jetson AGX Orin installed on the coral deployment vessel. In addition to describing the pipeline, we analyze the impact of the different components in terms of the operational requirements defined in Section~\ref{sec:operational}, and rank their performance accordingly in \tabref{table:ranking}.

\subsection{Image Labeling Scheme}
\label{subsec:imageannot}
The `Image Labeling Scheme' is the first component of our proposed pipeline. The choice of labeling scheme depends primarily on available annotation resources, including cost and access to domain experts, as well as on the level of information and accuracy required from the pipeline. In the following, we introduce and discuss the different labeling schemes:

\subsubsection{Human Expert Labeling}
\label{subsec:humanlabel}
Human expert labeling is the most expensive scheme, requiring both financial resources and expert marine scientists to annotate the data. Labeling can be performed at two levels: `Image-Level', where each image is assigned a single label, or `Patch-Level', where each image is divided into smaller crops that are individually labeled. Image-level labeling has a lower cost but generally yields lower accuracy for the overall pipeline compared to patch-level labeling. Furthermore, label noise is inevitable due to human errors and the inherent subjectivity of expert decisions.

\subsubsection{Patch-wise Pseudo-labeling with CLIP}
\label{subsec:pseudo}
Given the substantial time and cost demands of patch-level labeling, we instead leverage the human expert image-level labels, and then adopt the pseudo-labeling scheme from~\cite{raine2024image} to rapidly pseudo-label a large number of patches for training. The large vision-language model CLIP~\cite{radford2021learning}, as described in~\cite{raine2024image}, is capable of generating textual descriptions of a given image. CLIP has limited specialist knowledge, but its training on internet-scale data gives it the ability to reliably distinguish many common elements (\ie sand, water, coral). 

Patches matching the image-level label ``\textit{Coral}" are retained as ``\textit{Coral}", while patches identified as ``\textit{Sand}" or ``\textit{Water}" are labeled as ``\textit{No Deploy}".  This pseudo-labeling approach ensures reliable labeling for the images labeled as ``\textit{Coral}" and ``\textit{No Deploy}". For images labeled as ``\textit{Deploy}", any remaining patches that do not match either ``\textit{Coral}" or ``\textit{No Deploy}" are assigned to the ``\textit{Deploy}" class, accounting for the context-dependent nature of this category. Applying this pseudo-labeling scheme to 3,185 images resulted in 89,180 weakly labeled patches (28 per image) for training the patch-level classifier. This design choice reduces annotation cost considerably, at the expense of some decrease in overall system accuracy. 

\subsubsection{Unsupervised Patch-wise Labeling with ChatGPT-4o}
\label{subsec:chatgpt}
When data annotation is severely limited by time or resources, we propose using ChatGPT-4o~\cite{openai2025chatgpt4o} for completely unsupervised pseudo-labeling of seafloor patches. This labeling scheme enables new data to be quickly incorporated into training with minimal human input by leveraging the visual understanding capabilities of large multimodal models. While it drastically reduces annotation costs, the resulting classifier exhibits lower accuracy, reflecting the absence of domain-specific expert knowledge during labeling. We use careful prompt engineering for our ChatGPT pseudo-labeling and find the most effective prompt to be the one shown in \figref{fig:prompt}.
\begin{figure}[t]
\centering
\begin{minipage}{\linewidth}
    \begin{tcolorbox}[
        colback=blue!2,
        colframe=DarkPurple!100!,
        colbacktitle=DarkPurple!100!,
        title=\textbf{Prompt}: Underwater Image Classification Agent,
        arc=3mm,
        boxrule=0.8pt,
        top=1mm,
        bottom=1mm,
        left=2mm,
        right=2mm,
        boxsep=1mm,
        width=\linewidth
    ]
    \footnotesize
    You are a specialized agent in classifying underwater images. Your goal is to carefully inspect the image, and then classify it in one of the following classes: 0, 1, 2.

    \begin{itemize}
        \setlength{\itemsep}{0.2em}
        \setlength{\topsep}{0.2em}
        \item The class 1 corresponds to: mainly coral.
        \item The class 2 corresponds to: rocky seafloor or substrate. It looks solid and has minimal coral.
        \item The class 0 corresponds to: images that do not fit into class 1 or class 2, typically algae, sand, rubble, water or blurry images.
    \end{itemize}

    Pay attention to the image and only classify it as class 0 if you're absolutely certain the image cannot be described as class 1 or 2.

    Always provide the output as a dictionary in the format 
    \[\{class: 0,\:  conf: 0.5\} \] 
    where `class' is an integer number corresponding to the best class (0, 1, 2) and `conf' is a decimal number between 0 and 1 representing your confidence. If you think two classes could accurately describe the image, the confidence should be closer to 0.
    \end{tcolorbox}
\end{minipage}
\caption{Input prompt for the ChatGPT pseudo-labeling scheme.}
\label{fig:prompt}
\end{figure}

\subsection{Classifier}
The `Classifier' module is the pipeline component tasked with performing image analysis. In the following sections, we present the proposed classifiers and evaluate their advantages and limitations in the context of automated coral reseeding:
\subsubsection{Image-Level Classifier}
\label{subsec:imagetraining}
The `Image-Level Classifier' directly maps input images from the camera system to one of the output classes. This method is fast and requires minimal pre-processing, but its accuracy is limited. Images often contain mixed-class elements, and expert definitions of a \textit{suitable location} vary, introducing inconsistency. 

The image-level classification also provides limited feedback, making the interpretation of the model's output more difficult. The approach also lacks flexibility for multi-camera setups, as individual cameras may yield conflicting predictions, necessitating a separate voting mechanism. Training requires a fully labeled dataset from marine experts, but the labeling effort is confined to image-level categories. 

\begin{table}[t]
\caption{Ranking of different variations of the proposed pipeline based on operational requirements}
\label{table:ranking}
\centering
\small %
\renewcommand{\arraystretch}{1.2} %
\setlength{\tabcolsep}{3pt} %

\resizebox{\columnwidth}{!}{%
\begin{tabular}{c c c c c c c}
\toprule
\makecell{\textbf{Classifier}} & 
\makecell{\textbf{Labeling}} & 
\makecell{\textbf{Accuracy}} &
\makecell{\textbf{Inference}\\\textbf{Speed}} &
\makecell{\textbf{Interpretability}} &
\makecell{\textbf{Flexibility}} &
\makecell{\textbf{Labeling}\\\textbf{Efficiency}} \\
\midrule

\makecell[c]{\rotatebox[origin=c]{90}{\textbf{ Image }}} & Human 
& \mbox{\fulstar\fulstar\empstar} 
& \mbox{\fulstar\fulstar\fulstar} 
& \mbox{\empstar\empstar\empstar} 
& \mbox{\empstar\empstar\empstar} 
& \mbox{\fulstar\empstar\empstar} \\
\midrule

\makecell[c]{} & Human 
& \mbox{\fulstar\fulstar\fulstar} 
& \mbox{\fulstar\empstar\empstar} 
& \mbox{\fulstar\fulstar\fulstar} 
& \mbox{\fulstar\fulstar\empstar} 
& \mbox{\empstar\empstar\empstar} \\

\makecell[c]{\rotatebox[origin=c]{90}{\textbf{Patch}}} & Human + AI (CLIP) 
& \mbox{\fulstar\empstar\empstar} 
& \mbox{\fulstar\empstar\empstar} 
& \mbox{\fulstar\fulstar\empstar} 
& \mbox{\fulstar\fulstar\empstar} 
& \mbox{\fulstar\fulstar\empstar} \\

\makecell[c]{} & AI (ChatGPT) 
& \mbox{\fulstar\empstar\empstar} 
& \mbox{\fulstar\empstar\empstar} 
& \mbox{\fulstar\empstar\empstar} 
& \mbox{\fulstar\fulstar\empstar} 
& \mbox{\fulstar\fulstar\fulstar} \\

\bottomrule
\end{tabular}%
}
\end{table}

\subsubsection{Patch-Level Classifier}
\label{subsec:patchtraining}
The `Patch-Level Classifier' partitions the original image into a regular grid of smaller patches, with the model generating labels for each patch. This version of the classifier has the advantage that, given the high resolution of marine science images, processing smaller patches rather than resizing the entire image preserves more detail, enabling more informed decisions. Additionally, patches contain fewer mixed elements, which significantly improves classification accuracy.

The patch representation enables the ``\textit{Coral}" class to be used for estimating coral coverage. The Patch-Level Classifier also facilitates multi-camera integration (\ie cameras on the port and starboard sides), as classified patches from different images can be combined more directly into a unified decision.

Feedback to the operator is provided by reconstructing the original image from the classified patches, with each patch color-coded according to its predicted class, as shown in \figref{fig:frontpage}. This visualization gives operators insight into why the subsequent module will decide to deploy or not. 

The main limitations are that the model performs inference on multiple patches rather than a single image, reducing speed; and that patch-level labeling increases expert annotation burden, as described in the previous subsection. 

\subsection{Decision-Making Module}
\label{subsec:dispensing}
The `Decision-Making Module' depends on the selected classifier, either `Image-Level' or `Patch-Level'. In the `Image-Level' configuration, the decision process directly mirrors that of a human ecologist: the device is deployed if the predicted class is ``\textit{Deploy}", and withheld otherwise. Instead, the `Patch-Level' configuration requires an aggregation strategy to convert patch-level predictions into a single deployment decision. We propose two strategies.

\subsubsection{Thresholding with Patches}
The first strategy, `Thresholding with Patches', emulates the heuristic used by marine scientists by evaluating the proportion of deployable regions relative to non-deployable ones. In practice, we compute the ratio of the ``\textit{Deploy}" patches relative to the combined number of ``\textit{No-Deploy}" and ``\textit{Coral}" patches. Deployment occurs if this ratio exceeds a calibrated threshold $\alpha$.

\subsubsection{Spatial Patch Aggregation}
The second strategy, namely `Spatial Patch Aggregation', leverages a lightweight Convolutional Neural Network~\cite{9451544} to account not only for the deployable ratio but also for spatial relationships between neighboring patches. The network outputs a deployment probability, and the device is released when this probability surpasses the threshold $\alpha$.

\begin{table}[t]
\setlength{\tabcolsep}{1.3px}
\caption{Dataset summary and field trial locations} 
\label{table:sites}
\centering
\scriptsize
\begin{tabular}{>{\centering\arraybackslash}m{0.3cm} >{\centering\arraybackslash}m{1.7cm} >{\centering\arraybackslash}m{0.9cm} >{\centering\arraybackslash}m{1.75cm} >{\centering\arraybackslash}m{1.75cm} >{\centering\arraybackslash}m{1.75cm}}
\toprule
\textbf{Site} & \textbf{Location} & \textbf{Depth Range} & \textbf{Deployment Sequences} & \textbf{Patches} & \textbf{Weak Labels}\\
\midrule
1 & Combined Sites: Heron Island, Cairns, Moore Reef & 1.4-10.4m & \crossmark & \checkmark (2,191 patches) & \checkmark (601 images) \\
\arrayrulecolor{black!30}\midrule
2 & Maureen's Cove, Whitsundays & 1.9-6.4m & \checkmark (1,000 images) & \crossmark & \checkmark (1,043 images) \\
\midrule
3 & Black Island, Whitsundays & 2.2-5.8m & \checkmark (500 images) & \checkmark (1,944 patches) & \checkmark (582 images) \\
\midrule
4 & Unsafe Passage, Whitsundays & 1.4-8.3m & \checkmark (1,000 images) & \checkmark (3,000 patches) & \checkmark (359 images) \\
\midrule
5 & Heron Island & 1.5-8.5m & \checkmark (1,500 images) & \checkmark (3,000 patches) & \checkmark (600 images) \\
\arrayrulecolor{black!100}\bottomrule 
\end{tabular}
\end{table}

\subsection{Handling Class Imbalance}
\label{subsec:focal}
The collected datasets, described in \tabref{table:sites}, present a significant class imbalance, with a predominance of the ``\textit{No-Deploy}" category. To mitigate this issue, we adopt and extend the Focal Loss formulation~\cite{lin2017focal} to the multi-class classification scenario. The binary $w$-balanced Focal Loss is defined as:
\begin{align*}
\mathcal{L}_{\text{focal}} = &-w \sum_{i \in C_0}(1 - p_i)^{\gamma} \log(p_i)\\
&- (1 - w) \sum_{i \in C_1} p_{i}^{\gamma}\log(1 - p_i),
\end{align*}
where: 
  \( y_i \in \{ +1, -1 \} \) is the ground truth class label for sample \( i \). 
  The set \( C_0 \) contains samples labeled with \( y_i =+1 \), while \( C_1 \) contains samples labeled with \( y_i =-1 \). With a slight abuse of notation, we refer to the sets as the classes themselves. In the binary classification setting, the term \( p_i \) denotes the predicted probability for samples in class \( C_0 \), while the probability for samples in  \( C_1 \) is \( 1 - p_i \). \( w \) is a tunable weighting factor that balances the importance of positive and negative examples, and \( \gamma \) is the focusing parameter that reduces the contribution of easy examples while emphasizing hard ones.

To generalize this formulation to the multi-class case, let \( M \) denote the total number of classes such that \( y_i \in \{1, \ldots, M\} \). Let \( p_i \) the predicted probability for class \( C_k = \{y_i \mid y_i = k\} \). The total number of samples is then defined as:
\begin{equation*}
    N = \sum_{k=1}^{M} |C_k|,
\end{equation*}
where \( |C_k| \) is the cardinality of set \( C_k \). We compute the importance weight \( w_k \) for each class $C_k$ as the complement of their relative frequency:
\begin{equation*}
w_k = 1 - \frac{|C_k|}{N}. %
\end{equation*}
The final multi-class weighted focal loss is defined as:
\begin{equation*}
\mathcal{L}_{\text{focal}} = -\sum_{k=1}^{M} w_k \left(\sum_{i \in C_k}  (1 - p_{i})^{\gamma} \cdot \log(p_{i}) \right).
\end{equation*}

This loss enables the model to focus on hard-to-classify or underrepresented classes, while incorporating weighting to address dataset imbalance.

\section{Experimental Setup}
\label{sec:experimentalsetup}

\paragraph*{Implementation}

All experiments were completed on an NVIDIA A100 GPU.  Classifiers were implemented in PyTorch~\cite{paszke2019pytorch}, with ONNX used to convert the models for inference on the Jetson AGX Orin.  We initialized the models with weights pre-trained on ImageNet. To counter the class imbalance in the training data, we use weighted oversampling, data augmentation, and the multi-class version of the Focal Loss, defined in \secref{subsec:focal}, with $\gamma = 2.0$.

\paragraph*{Datasets}
The data used in this work was collected across four field trial locations on a large coral reef to maximize the diversity of environmental and visual conditions (Table~\ref{table:sites}).  Multiple labeling schemes were  employed, enabling training and testing for both patch and image-level inferences. There is no overlap between the data used for each annotation style.

The images are collected using a high-resolution optical imaging system mounted on a crewed surface vessel. 
In this work, we focus on operating conditions up to 10m depth from the surface (2-3m for inshore reefs with higher turbidity and 5-8m for mid-shelf reefs with relatively low turbidity).

\paragraph*{Evaluation}
We perform evaluation on two tasks: patch classification and deployment decisions. These tasks capture fine-grained resolution at the patch-level, and coarse-grained deployment accuracy at the image-level. 
For patch classification, we report the performance of our model on three classes: \textit{`Deploy'}, \textit{`No-Deploy'} and \textit{`Coral'}, using the per-class precision, recall, F1 scores, and the overall F1 score, which we calculate as \emph{macro} F1, \ie, the average of the per-class F1 scores.  This ensures that all classes are treated equally, which better evaluates the performance on class imbalanced datasets. For whole image classification, we instead focus on the \textit{`Deploy'} precision and recall, and the overall accuracy and F1 score. 
\begin{table}[t]
\setlength{\tabcolsep}{2px}
\caption{Performance and inference speeds for different variations of the Reef Guidance System}
\label{table:wholeframe}
\centering
\scriptsize
\begin{tabularx}{0.99\columnwidth}{@{}l>{\centering\arraybackslash}X>{\centering\arraybackslash}X>{\centering\arraybackslash}X>{\centering\arraybackslash}X>{\centering\arraybackslash}X>{\centering\arraybackslash}X}
\toprule
  & \textbf{Deploy} & \textbf{Deploy} & \textbf{Overall} & \textbf{Overall} & \textbf{Time}  \\
\textbf{Decision-making Module} & \textbf{Precision} & \textbf{Recall} & \textbf{Accuracy} & \textbf{F1} & \textbf{(s)} \\

\midrule
Image-Level Prediction & \textbf{65.21\%} & 38.02\% & 76.59\% & 48.03\% & 0.648 \\
\arrayrulecolor{black!30}\midrule
Spatial Patch Aggregation ($\alpha=0.3$) & 59.77\% & 67.16\%  & \textbf{77.78\%} & \textbf{63.25\%} & \textbf{0.209} \\
\arrayrulecolor{black!30}\midrule
Thresholding with Patches ($\alpha=0.4$) & 55.48\% & \textbf{70.24\%}  & 75.49\% & 61.99\% & 0.226 \\
\arrayrulecolor{black!100}\bottomrule 
\end{tabularx}
\end{table}

\captionsetup[subfloat]{font=small, labelfont=small, textfont=small}
\begin{figure*}[t]
    \centering
    \begin{minipage}[t]{0.52\textwidth}
        \centering
        \includegraphics[width=\linewidth, clip, trim=0.5cm 0.6cm 0.3cm 0.0cm]{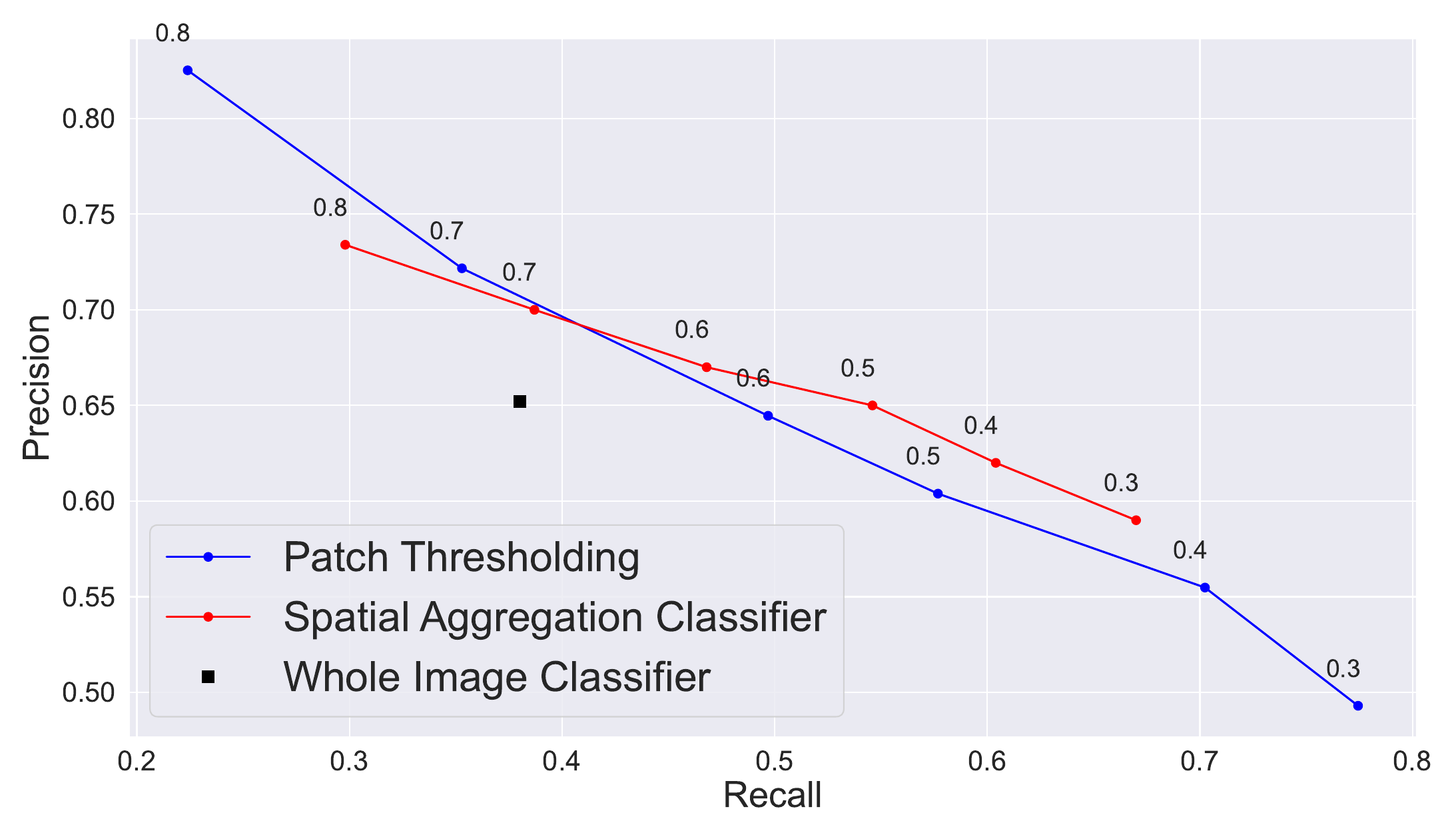}
        {\small (a) PR Curve}
    \end{minipage}
    \hfill
    \begin{minipage}[t]{0.42\textwidth}
        \centering
        \includegraphics[width=\linewidth, clip, trim=3cm 0cm 2.0cm 0cm]{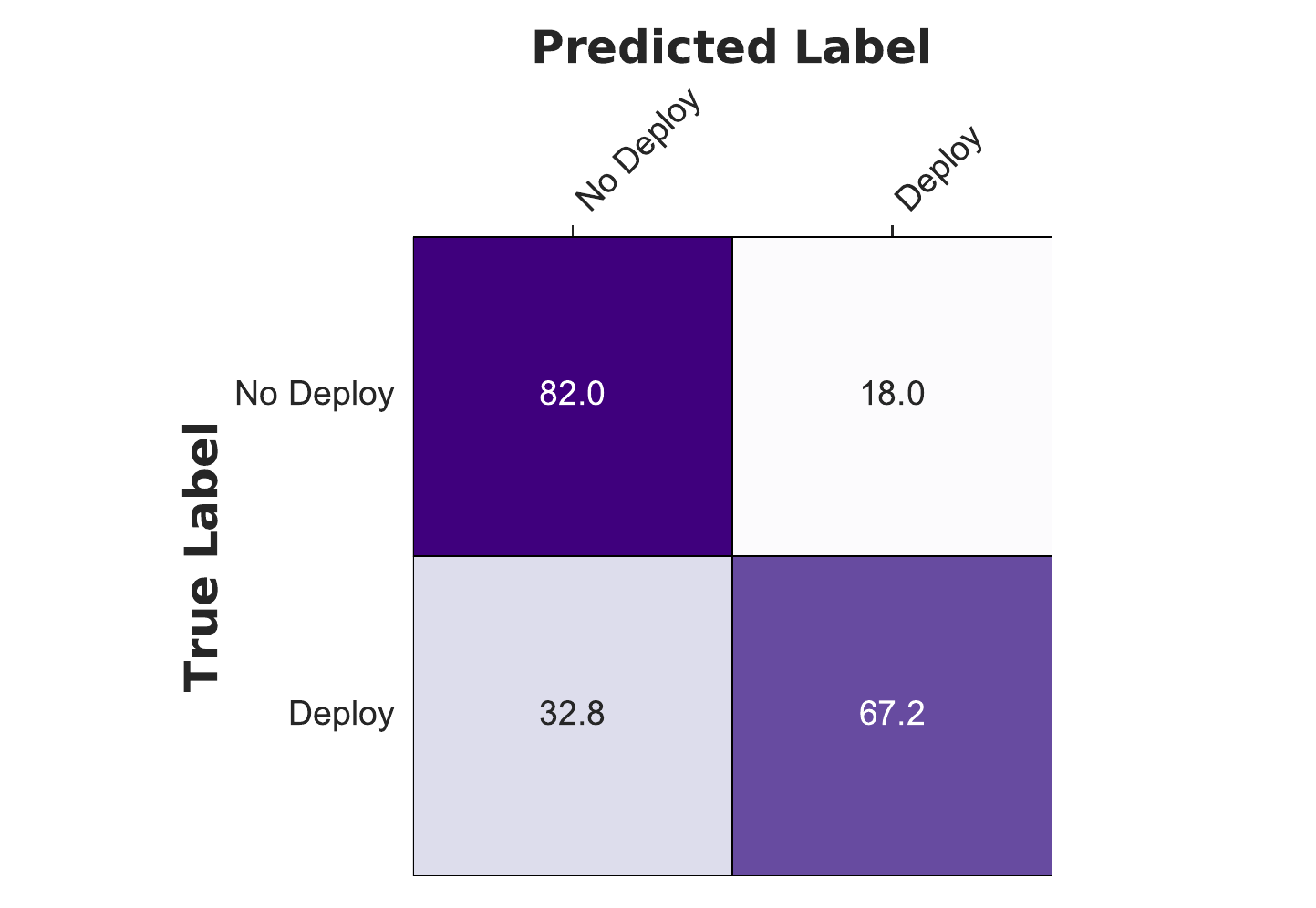}
        {\small (b) Confusion Matrix}
    \end{minipage}
    \caption{(a) Precision–Recall curve showing the different deployment approaches. The threshold $\alpha$ can be tuned based on operational requirements. A higher $\alpha$ increases recall for the \textit{Deploy} class, meaning that images labeled as \textit{No Deploy} are \textbf{less frequently} misclassified as \textit{Deploy}, but more \textit{Deploy} images are classified as \textit{No Deploy}. This corresponds to a more conservative deployment strategy. (b) Example confusion matrix for $\alpha=0.3$.}
    \label{fig:alpha}
\end{figure*}

\begin{table*}[t]
\setlength{\tabcolsep}{4.5px}
\vspace*{0.15cm}
\caption{Performance of different backbones and annotation schemes for Patch-Level Classification} 
\vspace*{-0.15cm}
\label{table:patchvsimage}
\centering
\scriptsize
\begin{tabular}{@{} lllcccccccccc @{}} %
 \toprule %
\multirow{2}[3]{*}{\textbf{Human Supervision}} & \multirow{2}[3]{*}{\textbf{Model Pseudo-labeling}}  & \multirow{2}[3]{*}{\textbf{Model}} &\multicolumn{3}{c}{\textbf{No Deploy}} & \multicolumn{3}{c}{\textbf{Coral}} & \multicolumn{3}{c}{\textbf{Deploy}} & \textbf{Overall} \\

 \cmidrule(lr){4-6} \cmidrule(lr){7-9} \cmidrule(lr){10-12}

& & & Prec. & Recall & F1 & Prec. & Recall & F1 & Prec. & Recall & F1 & \textbf{F1 Score} \\
\midrule
Patch labels & nil & MobileNet-v3-small~\cite{howard2019searching} & 95.39 & 92.14 & 93.74 & 81.43 & 87.25 & 84.24 &  82.34 & 86.20 & 84.22 & 87.40 \\
Patch labels & nil & MobileNet-v3-large~\cite{howard2019searching} & \textbf{95.81} & 93.18 & 94.47 & 84.22 & \textbf{88.52} & \textbf{86.32} & 83.29 & 86.98 & 85.10 & 88.63 \\
Patch labels & nil & EfficientNet-B0~\cite{tan2019efficientnet} & 95.69 & 93.84 & \textbf{94.76} & \textbf{85.60} & 84.95 & 85.28 & 84.26 & \textbf{90.63} & \textbf{87.33} & \textbf{89.12} \\
Patch labels & nil & ResNet-18~\cite{he2016deep} & 94.19 & \textbf{93.91} & 94.05 & 83.38 & 84.22 & 83.80 & \textbf{84.38} & 84.38 & 84.38 & 87.41 \\
\arrayrulecolor{black!100}\midrule
Image labels & CLIP~\cite{radford2021learning} & MobileNet-v3-small~\cite{howard2019searching} & 88.90 & 85.60 & 87.22 & 66.13
 & 73.03 & 69.41 & 62.60 & 64.06 & 63.32 & 73.32 \\
 nil & ChatGPT-4o~\cite{openai2025chatgpt4o} & MobileNet-v3-small~\cite{howard2019searching} & 83.89 & 81.53 & 82.69 & 81.73 & 62.60 & 70.89 &  50.49 & 67.62 & 57.81 & 70.47 \\
\arrayrulecolor{black!100}\bottomrule 
\end{tabular}
\vspace*{-0.3cm}
\end{table*}

\section{Results}
\label{sec:results}

We evaluate the proposed pipeline by comparing classification methods and foundation model pseudo-labeling (Section~\ref{subsec:methodcomparison}), ablation studies on feature extractors and deployment thresholds (Section~\ref{subsec:ablation}), and real-world field testing on the Great Barrier Reef alongside expert ecologists (Section~\ref{subsec:real}).

\subsection{Comparison of Proposed Approaches}
\label{subsec:methodcomparison}
Patch classification enables interpretation of model outputs, improving trust and transparency in the pipeline; and also enables sub-image deployment decisions if the operational setup enables coral device deployment from both sides of the vessel.  Patch-level classification outperforms image-level classification by 15.2\% overall F1 score (Table~\ref{table:wholeframe}).

We also investigated the performance of classifiers when trained with CLIP~\cite{radford2021learning} and ChatGPT-4o~\cite{openai2025chatgpt4o} pseudo-labeling. Table~\ref{table:patchvsimage} demonstrates that CLIP supervision, which requires only image labels, still achieves 73.32\% overall F1 score and ChatGPT-4o, which requires no human labels at all, achieves 70.47\% overall F1 score.  This highlights that large vision-language foundation models could be used to supplement human labels or as an additional source of pseudo-labeled data in scenarios where human annotation is not available or extremely limited.

\captionsetup[subfloat]{font=small, labelfont=small, textfont=small}
\begin{figure*}[t]
    \centering
    \begin{minipage}[t]{0.49\linewidth}
        \centering
        \includegraphics[width=\linewidth, clip, trim=8.5cm 3.0cm 8.5cm 3.0cm]{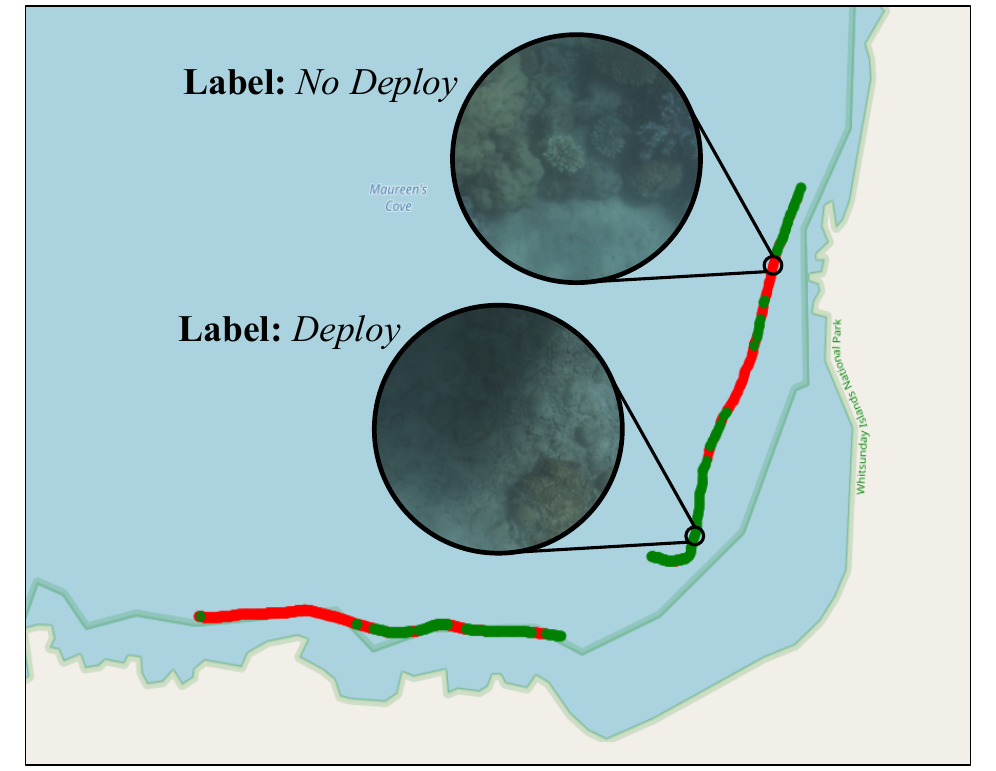}
        \vspace{2mm}
        {\small (a) Expert Marine Scientist Decisions}
    \end{minipage}
    \hfill
    \begin{minipage}[t]{0.49\textwidth}
        \centering
        \includegraphics[width=\linewidth, clip, trim=8.5cm 3.0cm 8.5cm 3.0cm]{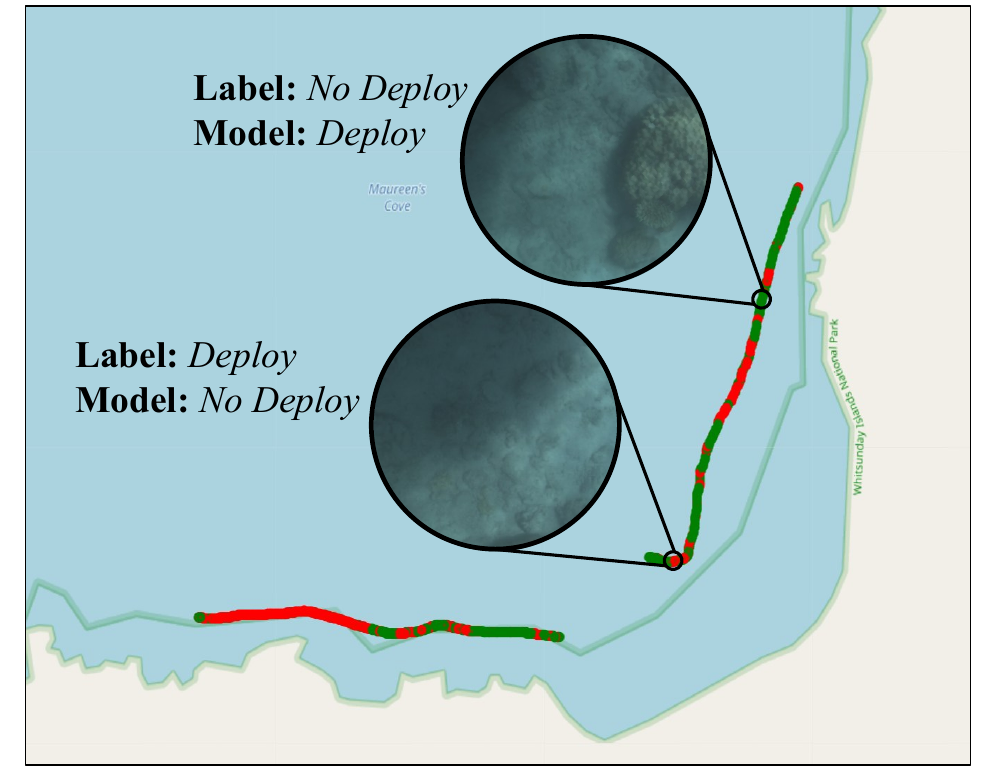}
        \vspace{2mm}
        {\small (b)  Proposed Pipeline Decisions}
    \end{minipage}
    \caption{Real-world deployment results for our proposed Reef Guidance System, visualized as GPS maps comparing an ecologist's deployment decisions (left) alongside the outputs of our proposed pipeline (right). Green markers indicate `\textit{Deploy}' frames, where coral devices can be released; red markers indicate `\textit{No Deploy}' frames where the corals should not be released.  Where the red and green markers are the same between the left and right images, this means both the ecologist and the pipeline are in agreement.  We highlight an example image for both `\textit{Deploy}' and `\textit{No Deploy}' on the left image, and also show incorrect decisions for our pipeline in the right image. These examples also show the difficulty of the images and semantic similarity between `\textit{Deploy}' and `\textit{No Deploy}' sample frames. Overall, our pipeline (right) matches closely with the ecologist's decisions.}
    \label{fig:realworld}
\end{figure*}

\subsection{Ablation Study}
\label{subsec:ablation}

We evaluate the impact of the backbone used for feature extraction, and we experiment with different values for the deployment threshold parameter, $\alpha$. 

\subsubsection{Feature Extractor}
\label{subsubsec:model}

Table~\ref{table:patchvsimage} evaluates various deep learning models in terms of their classification performance. One key operational requirement is real-time inference. MobileNet-v3 (`small' and `large')~\cite{howard2019searching}, ResNet-18~\cite{he2016deep} and EfficientNet-B0~\cite{tan2019efficientnet} perform inference at 0.226, 0.278, 0.258 and 0.346 seconds per image respectively.  Although we find that the highest performing backbone is the EfficientNet-B0, we select MobileNet-v3-small as the absolute difference in F1 score is only 1.7\% and it has the fastest inference speed.  Alternate backbones could be selected for different operational priorities.

\subsubsection{Deployment Threshold}
\label{subsubsec:threshold}

In the case of the image-level classifier, the predicted class `\textit{Deploy}' is used as the trigger for deploying the coral device. For the patch-level classifier, we find the deployment decision using either the `Spatial Patch Aggregation' or `Thresholding with Patches' strategies.  The deployment threshold $\alpha$ is the ratio of patches in an image classified as `\textit{Deploy}' versus the patches classified as `\textit{Coral}' or `\textit{No-Deploy}'; or in the case of the `Spatial Patch Aggregation', it is threshold we use to convert the output of our decision model to $0$ or $1$. The threshold can be tuned to make deployment more conservative (increase $\alpha$) if limited coral devices are available, or less conservative (decrease $\alpha$) if widespread deployment is desired. Varying $\alpha$ leads to the PR Curve shown in Fig.~\ref{fig:alpha}, which shows that both strategies are comparable, with the `Spatial Patch Aggregation' resulting in slightly higher performance and inference speeds.

\subsection{Real-world Field Testing}
\label{subsec:real}

We demonstrate our proposed pipeline on the Great Barrier Reef and evaluate the performance alongside an expert ecologist. The resulting deployment trajectories from these field trials are visualized in Fig.~\ref{fig:realworld}, alongside ecologist-provided deployment decisions. The model performed inference at 5.5 frames per second.  Fig.~\ref{fig:realworld} shows that the images are very difficult to classify, particularly in terms of the semantic similarity between images labeled as ``\textit{Deploy}'' and ``\textit{No-Deploy}".  In future work, we will investigate inter- and intra-observer variation, and explore impacts of annotator uncertainty on model performance. 
\section{Conclusion}
\label{sec:conc}

In this paper, we have practically implemented and demonstrated the first pipeline for broad-scale automated deployment of coral reseeding devices for restoration of the Great Barrier Reef. Our pipeline is flexible to the operational requirements: the required level of accuracy, availability of expert annotators, inference speed and the hardware setup \ie different vessels and coral deployment mechanisms.  We have shown how large vision-language models can be used effectively to pseudo-label data in annotation-constrained operational settings. We perform extensive field experiments and find that the accuracy of our AI-driven deployment system is comparable to an ecologist, while overcoming human limits and significantly increasing the potential for broad-scale coral reseeding. 

The performance of the classifier and deployment module is mainly constrained by the limited size of the training dataset and inherent label noise. Investigating inter-observer variability among annotators and integrating these insights into classifier training is a valuable direction for future work. Additionally, leveraging the `Patch-Level Classifier' for active path planning to identify optimal coral deployment sites represents another promising research avenue.
\bibliographystyle{IEEEtran} 
\bibliography{IEEEabrv,Bibliography} 

\begin{thebibliography}{10}
\providecommand{\url}[1]{#1}
\csname url@samestyle\endcsname
\providecommand{\newblock}{\relax}
\providecommand{\bibinfo}[2]{#2}
\providecommand{\BIBentrySTDinterwordspacing}{\spaceskip=0pt\relax}
\providecommand{\BIBentryALTinterwordstretchfactor}{4}
\providecommand{\BIBentryALTinterwordspacing}{\spaceskip=\fontdimen2\font plus
\BIBentryALTinterwordstretchfactor\fontdimen3\font minus \fontdimen4\font\relax}
\providecommand{\BIBforeignlanguage}[2]{{%
\expandafter\ifx\csname l@#1\endcsname\relax
\typeout{** WARNING: IEEEtran.bst: No hyphenation pattern has been}%
\typeout{** loaded for the language `#1'. Using the pattern for}%
\typeout{** the default language instead.}%
\else
\language=\csname l@#1\endcsname
\fi
#2}}
\providecommand{\BIBdecl}{\relax}
\BIBdecl

\bibitem{henley2024highest}
B.~J. Henley, H.~V. McGregor, A.~D. King, O.~Hoegh-Guldberg, A.~K. Arzey, D.~J. Karoly, J.~M. Lough, T.~M. DeCarlo, and B.~K. Linsley, ``Highest ocean heat in four centuries places great barrier reef in danger,'' \emph{Nature}, vol. 632, no. 8024, pp. 320--326, 2024.

\bibitem{hughes2023principles}
T.~P. Hughes, A.~H. Baird, T.~H. Morrison, and G.~Torda, ``Principles for coral reef restoration in the anthropocene,'' \emph{One Earth}, vol.~6, no.~6, pp. 656--665, 2023.

\bibitem{randall2020sexual}
C.~J. Randall, A.~P. Negri, K.~M. Quigley, T.~Foster, G.~F. Ricardo, N.~S. Webster, L.~K. Bay, P.~L. Harrison, R.~C. Babcock, and A.~J. Heyward, ``Sexual production of corals for reef restoration in the {Anthropocene},'' \emph{Marine Ecology Progress Series}, vol. 635, pp. 203--232, 2020.

\bibitem{banaszak2023applying}
A.~T. Banaszak, K.~L. Marhaver, M.~W. Miller, A.~C. Hartmann, R.~Albright, M.~Hagedorn, P.~L. Harrison, K.~R. Latijnhouwers, S.~Mendoza~Quiroz, V.~Pizarro \emph{et~al.}, ``Applying coral breeding to reef restoration: best practices, knowledge gaps, and priority actions in a rapidly-evolving field,'' \emph{Restoration Ecology}, vol.~31, no.~7, p. e13913, 2023.

\bibitem{tsai2026automated}
D.~Tsai, C.~A. Brunner, R.~Lamont, F.~M. Nordborg, A.~Severati, J.~Terry, K.~Jackel, M.~Dunbabin, T.~Fischer, and S.~Raine, ``Automated coral spawn monitoring for reef restoration: The coral spawn and larvae imaging camera system (cslics),'' in \emph{IEEE International Conference on Robotics and Automation}, 2026.

\bibitem{severati2024autospawner}
A.~Severati, F.~M. Nordborg, A.~Heyward, M.~A.~A. Wahab, C.~A. Brunner, J.~Montalvo-Proano, and A.~P. Negri, ``The autospawner system-automated ex situ spawning and fertilisation of corals for reef restoration,'' \emph{Journal of Environmental Management}, vol. 366, p. 121886, 2024.

\bibitem{gibbs2021technology}
M.~T. Gibbs, ``Technology requirements, and social impacts of technology for at-scale coral reef restoration,'' \emph{Technology in Society}, vol.~66, p. 101622, 2021.

\bibitem{raine2026ai}
S.~Raine and T.~Fischer, ``{AI-driven marine robotics: E}merging trends in underwater perception and ecosystem monitoring,'' in \emph{Proceedings of the AAAI Conference on Artificial Intelligence}, 2026.

\bibitem{belcher2023demystifying}
B.~T. Belcher, E.~H. Bower, B.~Burford, M.~R. Celis, A.~K. Fahimipour, I.~L. Guevara, K.~Katija, Z.~Khokhar, A.~Manjunath, S.~Nelson \emph{et~al.}, ``Demystifying image-based machine learning: a practical guide to automated analysis of field imagery using modern machine learning tools,'' \emph{Frontiers in Marine Science}, vol.~10, p. 1157370, 2023.

\bibitem{raine2024reducing}
S.~Raine, F.~Maire, N.~Suenderhauf, and T.~Fischer, ``Reducing label dependency for underwater scene understanding: A survey of datasets, techniques and applications,'' \emph{arXiv preprint arXiv:2411.11287}, 2024.

\bibitem{huang2024research}
D.~Huang, Y.~Sun, W.~Gao, W.~Xu, W.~Wang, Y.~Zhang, and L.~Wang, ``Research on seamount substrate classification method based on machine learning,'' \emph{Frontiers in Marine Science}, vol.~11, p. 1431688, 2024.

\bibitem{ditria2022artificial}
E.~M. Ditria, C.~A. Buelow, M.~Gonzalez-Rivero, and R.~M. Connolly, ``Artificial intelligence and automated monitoring for assisting conservation of marine ecosystems: A perspective,'' \emph{Frontiers in Marine Science}, vol.~9, p. 918104, 2022.

\bibitem{kumar2025harnessing}
M.~R. Kumar and K.~Neppolian, ``Harnessing artificial intelligence and remote sensing for large-scale marine habitat monitoring and conservation,'' \emph{Journal of Animal Environment}, vol.~17, no.~3, pp. 391--402, 2025.

\bibitem{darkhal2025optimizing}
H.~Darkhal, K.~Kabiri, and M.~R. Shokri, ``Optimizing coral reef restoration: A gis-based site selection for coral nurseries in kish island, the persian gulf,'' \emph{Journal for Nature Conservation}, vol.~86, p. 126898, 2025.

\bibitem{zheng2024coralscop}
Z.~Zheng, H.~Liang, B.-S. Hua, Y.~H. Wong, P.~Ang, A.~P.~Y. Chui, and S.-K. Yeung, ``Coralscop: segment any coral image on this planet,'' in \emph{Proceedings of the IEEE/CVF Conference on Computer Vision and Pattern Recognition}, 2024, pp. 28\,170--28\,180.

\bibitem{raine2022point}
S.~Raine, R.~Marchant, B.~Kusy, F.~Maire, and T.~Fischer, ``Point label aware superpixels for multi-species segmentation of underwater imagery,'' \emph{IEEE Robotics and Automation Letters}, vol.~7, no.~3, pp. 8291--8298, 2022.

\bibitem{raine2024human}
S.~Raine, R.~Marchant, B.~Kusy, F.~Maire, N.~Sunderhauf, and T.~Fischer, ``Human-in-the-loop segmentation of multi-species coral imagery,'' in \emph{Proceedings of the IEEE/CVF Conference on Computer Vision and Pattern Recognition Workshops}, 2024, pp. 2723--2732.

\bibitem{raine2024image}
S.~Raine, R.~Marchant, B.~Kusy, F.~Maire, and T.~Fischer, ``Image labels are all you need for coarse seagrass segmentation,'' in \emph{Proceedings of the IEEE/CVF Winter Conference on Applications of Computer Vision}, 2024, pp. 5943--5952.

\bibitem{noman2023improving}
M.~K. Noman, S.~M.~S. Islam, J.~Abu-Khalaf, S.~M.~J. Jalali, and P.~Lavery, ``Improving accuracy and efficiency in seagrass detection using state-of-the-art {AI} techniques,'' \emph{Ecological Informatics}, vol.~76, 2023.

\bibitem{zhang2024point}
M.~Zhang, R.~Santa~Cruz, Y.~Arzhaeva, X.~Li, B.~Do, J.~Oorloff, M.~A. Armin, Z.~Hayder, and D.~Ahmedt-Aristizabal, ``Point-supervised seagrass segmentation for {3D} underwater habitat mapping,'' in \emph{2024 International Conference on Digital Image Computing: Techniques and Applications (DICTA)}.\hskip 1em plus 0.5em minus 0.4em\relax IEEE, 2024, pp. 54--61.

\bibitem{chen2021new}
Q.~Chen, O.~Beijbom, S.~Chan, J.~Bouwmeester, and D.~Kriegman, ``A new deep learning engine for coralnet,'' in \emph{Proceedings of the IEEE/CVF international conference on computer vision}, 2021, pp. 3693--3702.

\bibitem{gonzalez2020monitoring}
M.~Gonzalez-Rivero, O.~Beijbom, A.~Rodriguez-Ramirez, D.~E. Bryant, A.~Ganase, Y.~Gonzalez-Marrero, A.~Herrera-Reveles, E.~V. Kennedy, C.~J. Kim, S.~Lopez-Marcano \emph{et~al.}, ``Monitoring of coral reefs using artificial intelligence: A feasible and cost-effective approach,'' \emph{Remote Sensing}, vol.~12, no.~3, p. 489, 2020.

\bibitem{wyatt2022using}
M.~Wyatt, B.~Radford, N.~Callow, M.~Bennamoun, and S.~Hickey, ``Using ensemble methods to improve the robustness of deep learning for image classification in marine environments,'' \emph{Methods in Ecology and Evolution}, vol.~13, no.~6, pp. 1317--1328, 2022.

\bibitem{jagadeesh2025coral}
M.~Jagadeesh and U.~Pradhan, ``Coral reef bleaching prediction: A machine learning approach using environmental factors,'' in \emph{2025 3rd International Conference on Sustainable Computing and Data Communication Systems (ICSCDS)}.\hskip 1em plus 0.5em minus 0.4em\relax IEEE, 2025, pp. 2155--2160.

\bibitem{hanke2025out}
G.~Hanke, M.~Canals, R.~Nakajima, M.~Bergmann, F.~Galgani, D.~Li, G.~Papatheodorou, C.~K. Pham, D.~J. Amon, M.~Angiolillo \emph{et~al.}, ``Out of sight, but not out of mind: Key issues regarding seafloor macrolitter monitoring: Issued by the expert community “international seafloor macrolitter imaging and quantification”,'' \emph{Marine pollution bulletin}, vol. 221, p. 118500, 2025.

\bibitem{wyatt2026signal}
M.~Wyatt, J.~Vercelloni, C.~Faubel, J.~Colquhuon, M.~Wakeford, S.~Wilson, and C.~J. Fulton, ``Signal or noise? minimising errors in image-based ai for marine ecosystem monitoring,'' \emph{Landscape Ecology}, vol.~41, no.~1, p.~16, 2026.

\bibitem{jackett2023benthic}
C.~Jackett, F.~Althaus, K.~Maguire, M.~Farazi, B.~Scoulding, C.~Untiedt, T.~Ryan, P.~Shanks, P.~Brodie, and A.~Williams, ``A benthic substrate classification method for seabed images using deep learning: Application to management of deep-sea coral reefs,'' \emph{Journal of Applied Ecology}, vol.~60, no.~7, 2023.

\bibitem{loureiro2024survey}
G.~Loureiro, A.~Dias, J.~Almeida, A.~Martins, S.~Hong, and E.~Silva, ``A survey of seafloor characterization and mapping techniques,'' \emph{Remote Sensing}, vol.~16, no.~7, p. 1163, 2024.

\bibitem{morand2022identifying}
G.~Morand, S.~Dixon, and T.~Le~Berre, ``Identifying key factors for coral survival in reef restoration projects using deep learning,'' \emph{Aquatic Conservation: Marine and Freshwater Ecosystems}, vol.~32, no.~11, pp. 1758--1773, 2022.

\bibitem{aguzzi2024new}
J.~Aguzzi, L.~Thomsen, S.~Fl{\"o}gel, N.~J. Robinson, G.~Picardi, D.~Chatzievangelou, N.~Bahamon, S.~Stefanni, J.~Griny{\'o}, E.~Fanelli \emph{et~al.}, ``New technologies for monitoring and upscaling marine ecosystem restoration in deep-sea environments,'' \emph{Engineering}, 2024.

\bibitem{dunbabin2020uncrewed}
M.~Dunbabin, J.~Manley, and P.~L. Harrison, ``Uncrewed maritime systems for coral reef conservation,'' in \emph{Global Oceans 2020: Singapore--US Gulf Coast}.\hskip 1em plus 0.5em minus 0.4em\relax IEEE, 2020, pp. 1--6.

\bibitem{radford2021learning}
A.~Radford, J.~W. Kim, C.~Hallacy, A.~Ramesh, G.~Goh, S.~Agarwal, G.~Sastry, A.~Askell, P.~Mishkin, J.~Clark \emph{et~al.}, ``Learning transferable visual models from natural language supervision,'' in \emph{International Conference on Machine Learning}, 2021, pp. 8748--8763.

\bibitem{openai2025chatgpt4o}
OpenAI, ``Chatgpt (gpt-4o),'' \url{https://chat.openai.com/chat}, 2025, accessed: Jul. 10, 2025.

\bibitem{9451544}
Z.~Li, F.~Liu, W.~Yang, S.~Peng, and J.~Zhou, ``A survey of convolutional neural networks: Analysis, applications, and prospects,'' \emph{IEEE Transactions on Neural Networks and Learning Systems}, vol.~33, no.~12, pp. 6999--7019, 2022.

\bibitem{lin2017focal}
T.-Y. Lin, P.~Goyal, R.~Girshick, K.~He, and P.~Doll{\'a}r, ``Focal loss for dense object detection,'' in \emph{Proceedings of the IEEE international conference on computer vision}, 2017, pp. 2980--2988.

\bibitem{paszke2019pytorch}
A.~Paszke \emph{et~al.}, ``{PyTorch: An} imperative style, high-performance deep learning library,'' in \emph{Neural Information Processing Systems}, 2019.

\bibitem{howard2019searching}
A.~Howard, M.~Sandler, G.~Chu, L.-C. Chen, B.~Chen, M.~Tan, W.~Wang, Y.~Zhu, R.~Pang, V.~Vasudevan \emph{et~al.}, ``Searching for mobilenetv3,'' in \emph{Proceedings of the IEEE/CVF international conference on computer vision}, 2019, pp. 1314--1324.

\bibitem{tan2019efficientnet}
M.~Tan and Q.~Le, ``Efficientnet: Rethinking model scaling for convolutional neural networks,'' in \emph{International conference on machine learning}.\hskip 1em plus 0.5em minus 0.4em\relax PMLR, 2019, pp. 6105--6114.

\bibitem{he2016deep}
K.~He, X.~Zhang, S.~Ren, and J.~Sun, ``Deep residual learning for image recognition,'' in \emph{Proceedings of the IEEE conference on computer vision and pattern recognition}, 2016, pp. 770--778.

\end{thebibliography}

\end{document}